# IMPROVING THE PERFORMANCE OF THE RIPPER IN INSURANCE RISK CLASSIFICATION : A COMPARITIVE STUDY USING FEATURE SELECTION

Mlungisi Duma, Bhekisipho Twala, Tshilidzi Marwala
*Department of Electrical Engineering and the Built Environment, University of Johannesburg APK, Corner Kingsway and University Road, Auckland Park, South Africa*
mlungisi.duma@multichoice.co.za, btwala@uj.ac.za, tmarwala@uj.ac.za

Fulufhelo V. Nelwamondo
*Council for Scientific and Industrial Research (CSIR),
Pretoria, South Africa*
fnelwamondo@csir.co.za

Abstract: The Ripper algorithm is designed to generate rule sets for large datasets with many features. However, it was shown that the algorithm struggles with classification performance in the presence of missing data. The algorithm struggles to classify instances when the quality of the data deteriorates as a result of increasing missing data. In this paper, a feature selection technique is used to help improve the classification performance of the Ripper model. Principal component analysis and evidence automatic relevance determination techniques are used to improve the performance. A comparison is done to see which technique helps the algorithm improve the most. Training datasets with completely observable data were used to construct the model and testing datasets with missing values were used for measuring accuracy. The results showed that principal component analysis is a better feature selection for the Ripper in improving the classification performance.

## 1 INTRODUCTION

Insurance companies have played vital role in carrying risks on behalf of customers for years. These include the risk of covering the cost of a motor vehicle in case a client becomes involved in an accident. Another includes the risk of covering the costs for a client that was admitted to a hospital. However, a large number of people are still without insurance for a number of reasons. The first reason is affordability (Howe, 2010). The premiums for a cover maybe expensive to pay, therefore a customer is left with a choice of cancelling. The second reason is cancellation. A lotof claims or committing fraud by a customer results in their policy being terminated by the insurer. The third reason is refusing to get an insurance cover (Howe, 2010, Crump, 2009). Some people may feel that they can save enough money to cover the risk if something serious happens to them (Crump, 2009).

In this paper, we present a solution to improve the Ripper algorithm as a predictive modelling technique. The solution improves the way it predicts customer behaviour using past data in the insurance domain. The model learns past data about customers who are likely to have insurance cover (the data consists of a large number of attributes). This information is then used to predict the future behaviour of a different customer. A different customer data in this case has attributes with missing data. Missing because the data either not supplied by the customer or processing error by the system handling the data (Duma *et al.*, 2010).

In comparison with other supervised learning algorithms, the Ripper algorithm struggles with classification performance if the new data contain attributes missing data (Duma *et al.*, 2010). The main reason is over-fitting. The algorithm learns too much detail about the attributes of the customer data. The consequence of this was less accuracy in predicting new customer data. The accuracy is further impacted when the quality of the new data decreases as a result of increasing missing data. The algorithm showed less resilience in the presence of increasing missing data. This resulted in poor classification performance compared to other supervised algorithms such as the naïve Bayes, k-Nearest Neighbour, support vector machines and the logical discriminant analysis algorithm (Duma *et al.*, 2010).

We propose feature selection as a technique to improve the classification performance of the Ripper algorithm. Feature selection technique removes those attribute that are irrelevant. There are two feature selection techniques used in this paper, namely the principal component analysis and automatic relevance determination techniques .These techniques were selected primary because they are very effective data analysis and reduction techniques in high dimensional spaces.

Principal component analysis has been used in conjunction with classification algorithms to successfully identify cancer molecular patterns in micro-array data (Han, 2010). It has also been used as a feature selection technique in the automatic classification of ultra-sound liver images (Balasubramanian *et al.*, 2007) and in fault identification and analysis of vibration data (Marwala,2001). Automatic relevance determination technique has been applied successfully in selecting the most relevant features for classifying ovarian tumors (van Calster *et al.*, 2006). It has also been utilised successfully in the classification of myo-cardinal ischaema (a heart problem that occurs when there is lack of oxygen and nutrients, which results in arrhythmias and myocardial infractions) events (Smyrnakis *et al.*, 2007).

In this paper, feature selection technique removes those attribute that are irrelevant for classification. The remaining attributes are passed on to the Ripper algorithm to learn. This reduces over-fitting by the algorithm because it has less attributes to learn from. The result is more generality and increase in resilience, which results in improved accuracy.

The rest of this paper is organised as follows: Section 2 discusses the theoretical background on the Ripper, principal component analysis, automatic relevance determination and concludes with a discussion on missing data mechanisms. Section 3 is a discussion on the dataset and pre-processing, the PCA-Rip structure, as well as the ARD-Rip structure. Section 4 is a discussion on the experimental results. Section 5 gives a conclusion to the paper

## 2 BACKGROUND

## 2.1 RIPPER ALGORITHM

The Repeated Incremental Pruning to Produce Error Reduction (Ripper) is a classification algorithm designed to generate rules set directly from the training dataset. The name is drawn from the fact that the rules are learned incrementally. A new rule associated with a class value will cover various attributes of that class .The algorithm was designed to be fast and effective when dealing with large and noisy datasets compared to decision trees (Cohen, 1995).

The Ripper algorithm is illustrated by Algorithm 1 (adapted from (Cohen, 1995)):

---

**Algorithm 1:** Ripper Algorithm

**Input** : Training dataset S with *n* instances and *m* attributes

**Output** : Ruleset

**begin**
  sort classes in the order of least prevalent class to the most prevalent class.
  create a new rule set
    **while** iterating from the prevalent class to the most prevalent class
      split S into into $S_{pos}$ and $S_{neg}$
    **while** $S_{pos}$ is not empty
    split $S_{Pos}$ and $S_{neg}$ into $G_{pos}$ and $G_{neg}$ subsets and $P_{pos}$ and $P_{neg}$ subsets.
    create and prune a new rule
      **if** the error rate of the new rule is very large **then**
        **end while**
      **else**
        add new rule to rule set
        the total description length *l* is computed
      **if** $l > d$ **then**
        **end while**
    **end while**
  **end while**
**end**

---

1. S = {X, C} represents the training set, where X = {$x_1, x_2,…,x_k$} $\in \mathbb{R}^d$ represents the instances and C = {c1,c2,…,ck} $\in \mathbb{Z}$ represents the class-label associated with each instance.

2. The classes $c_1,...,c_k$ are sorted in the order of least prevalent class to the most frequent class. This is done by counting the number instances associated with each class. The instances associated with the least prevalent class are separated into $S_{Pos}$ subset whilst the remaining instances are grouped into $S_{neg}$ subset.
3. IREP is invoked (with $S_{Pos}$ and $S_{neg}$ subsets passed as parameters) to find the rule set that splits least prevalent class from the other classes.
4. Initialise an empty rule set R.
5. $S_{Pos}$ and $S_{neg}$ are split into growing positive $G_{pos}$ and growing negative $G_{neg}$ subsets as well as pruning positive $P_{pos}$ and negative $P_{neg}$ subsets. Growing positive subsets contains instances that are associated with the least prevalent class. Growing negative subset contains instances associated with the remaining classes. This is similar to the $P_{pos}$ and $P_{neg}$ subsets.
6. A new rule is created by growing $G_{pos}$ and $G_{neg}$. This is done by iteratively adding conditions that maximize the information gain criterion until the rule cannot cover any negative instances from the growing dataset.
7. The new rule is pruned for optimization of the function

$$v = \frac{p-n}{p+n} \quad (1)$$

using $P_{pos}$ and $P_{neg}$ subsets. *p* is number of rules to prune and *n* is the total number of rules.
8. Check the error rate of the new rule very large, and then return the rule set. Otherwise, the new rule is added to the rule set and the total *description length* is computed. If the lengths exceeds a certain number *d*, then the algorithm stops, otherwise repeat from step 5.
9. Iterate to the next least prevalent class and then repeat from step 3.

During the growing phase of the algorithm, a greedy approach of learning is applied, i.e. each rule is learned one at a time. In datasets with very large dimensions, this causes over-fitting of the data. This in turn increases the classification error rate significantly if the algorithm is tested with data with missing values.

The Ripper model is not as popular as the decision trees in the insurance domain, but it has been applied in financial risk analysis. It has been used in financial institutes to help find the best policy for credit products, increase revenue as well as decreasing losses (Peng, 2008).

## 2.2 PRINCIPAL COMPONENT ANALYSIS

Principal component analysis (PCA) is a feature selection technique used for pattern recognition in data with high dimensions (Marwala, 2009). The data can be represented in ways that can be used to express the similarities and differences. Furthermore, the data can be compressed into lower dimensional spaces.

In the majority of cases, the objective of the principal component analysis is to reduce the dimensions of the data whilst preserving as much as possible the representation of the original data. To achieve this, the initial step is to calculate the mean of each dimension and then subtract from the data. Thereafter, the covariance matrix of the data set is calculated. The eigenvalues as well as the eigenvectors are calculated using the covariance matrix as a basis. At this point, any vector dimension or its mean can be expressed as a linear combination of the eigenvectors. The final step is to choose the highest eigenvalues that correspond to the largest eigenvectors, known as the *principal components*. This step is where the idea of data compression comes to effect. The chosen eigenvalues along with their corresponding eigenvectors are used to reduce the dimensions without much loss of information (Marwala, 2009). This reduction can be expressed as

$$[T] = [A] \times [B] \quad (2)$$

where [T] is the transformed data set, [A] is the given data set and [B] is the principal component matrix. [T] represents a dataset that expresses the relationships between the data regardless of whether the data has equal or lower dimension. The original data set can be calculated using the following equation

$$[A'] = [T] \times [B^{-1}] \quad (3)$$

where [A'] is the re-transformed data set and [A'] ≈ [A] if all the data from $[B^{-1}]$ is used from the covariance matrix.

## 2.3 BAYESIAN ARTIFICIAL NEURAL NETWORK

Bayesian artificial neural network is a classifier that combines artificial neural network and Bayes theorem using probability distribution (Bishop, 1995). Suppose we have a two-layered artificial neural network with $\mathbf{x} \in \mathbb{R}^d$ input vectors in the input layer, *n* hidden layers, and a target value $t \in \{0,1\}$ in the output layer. The network can be expressed in the form

$$y_k = f(\sum_{j=0}^{M} w_{kj} g(\sum_{i=0}^{N} w_{ji} x_i)) \quad (4)$$

where $y_k$ is the output of the artificial neural network, *f* is the activation function from the hidden layer to the output layer, $w_{kj}$ are the weights from the j$^{th}$ hidden input connected to the k$^{th}$ output unit. The function *g* is the sigmoid activation from the input layer to the output layer, $w_{ji}$ are weights from the i$^{th}$ input unit connected to the j$^{th}$ hidden unit. The output $y_k$ is expressed as the posterior probability $P(y=1/\mathbf{x})$ and $P(y=0/\mathbf{x}) = 1 - y_k$.

The artificial neural network is trained using a dataset $D = \{\mathbf{x}, \mathbf{t}\}$ by iteratively adjusting the weights so as to minimize the log-likelihood error function (or the objective function) $E_D(\mathbf{w})$. The minimization is based on the continuous re-evaluation of the gradient of $E_D$ using the *back-propagation* technique. If a weight decay function $E_w = \frac{1}{2} \sum_i w_i^2$ is added to $E_D(\mathbf{w})$, the objective function changes to

$$F(\mathbf{w}) = -E_D(\mathbf{w}) + \alpha E_w \quad (5)$$

where $\alpha$ is the alpha hyper-parameter. The term $\alpha E_w$ regularizes the weight vector by penalizing weights with larger values to keep the neural network from over-fitting. The evidence approach to Bayesian modeling is to find the optimal (or most probable) values for $\alpha = \alpha_{MP}$ rather than integrating over them. This can be obtained from the equation, (MacKay, 1995)

$$\alpha_{MP} = 1 \Big/ \sum_i \frac{\gamma}{w_i^{MP}} \quad (6)$$

where $\gamma = k - trace(\mathbf{\Sigma}^{-1})$, *k* is the total number of parameters and $\mathbf{\Sigma}^{-1}$ is the variance - covariance matrix that defines the error bars on $\mathbf{w}$ parameters.

Using the hyper-parameters $\alpha_{MP}$ the optimal weights $\mathbf{w}_{MP}$ are determined by approximating the posterior $P(\mathbf{w}_{MP}|\alpha_{MP})$ by a Gaussian density expressed in the form

$$P(\mathbf{w}_{MP}|\alpha_{MP}) = \frac{1}{Z_s} \exp(-G - \alpha E_w) \quad (7)$$

where *G* is the cross-entropy function and $Z_s$ is the normalization constant.

## 2.4 AUTOMATIC RELEVANCE DETERMINATION

Automatic Relevance Determination (ARD) is a technique that is uses Bayes inference to identify and remove attributes that are not relevant to the prediction of the output variable (Mackay, 1995). This is achieved by assigning the hyper-parameter $\alpha_k$ to a group of weights that connect from the i$^{th}$ input variable. In a two-layered artificial neural network, each hyper-parameter is assigned to a group of weights connecting i$^{th}$ input to the hidden outputs, and from the j$^{th}$ hidden unit to the output units.

The hyper-parameter becomes large if the input is irrelevant, preventing them from causing major over-fitting. Using a Gaussian expression, the prior probability for each weight given $\alpha_k$ for each class, can be expressed as

$$P(w_i|\alpha_k) = \frac{1}{\sqrt{2\pi\alpha}} \exp(-\sum_k \alpha_k (\sum_{i \in k} w_i^2/2)) \quad (8)$$

Once the artificial neural network has been trained, the hyper-parameters are optimized using the evidence framework. The evidence finds the most probable value $\hat{\alpha}_k > 0$ which must satisfy equation (6).

## 2.4 MISSING DATA MECHANISMS

There are a number of reasons why data collected can have miss data. Well-known reasons include faulty processing by a system handling the data, different systems communicating with each other missing information or clients refusing to disclose all their information (Francis, 2005). It is imperative to know the reasons for data missing. When that reason is known, then appropriate methods for handling missing data are selected, which results in high prediction or classification accuracy.

The missing data mechanisms found currently in literature are *missing at random*, *missing completely*

at random, *missing not at random* and *missing by natural design* (Little *et al*, 1987, Marwala, 2009). Missing at random is a situation where the missing data is not related to the missing variables themselves but on other variables. Missing complete at random implies that the missing data is not dependent on any other existing data. Missing not at random implies a situation where the missing data depends on itself and not any other variables (Little *et al*, 1987, Marwala, 2009). Missing by natural design occurs when there is data missing because the variable is naturally deemed un-measurable, even though they are useful for analysis. In this case, the missing values are modelled using mathematical techniques (Marwala, 2009).

In this paper, we presuppose that the data is missing completely at for the problem under discussion. It is chosen so that single and multiple imputations return unbiased estimates.

## 3 METHODS

### 3.1 DATASETS AND PRE-PROCESSING

The experiment was conducted using two insurance datasets. The first insurance dataset was obtained from the University of California Irvine (UCI) machine learning repository. The dataset is used to predict which customers are likely to have an interest in buying a caravan insurance policy. In this paper, we are interested in finding out customers who are likely to have a car insurance policy, provided there is missing information.

The training dataset consists of over 5400 instances of which 5000 were used for the experiment. The testing dataset consists of only 4000 instances. Each set has a total of 86 attributes with completely observable data, 5 of which are categorical numeric values and 80 are continuous numeric values. The class attribute consists of only two values (0 to indicate a customer that is likely not to have insurance or 1 to indicate a customer that is likely to have an insurance cover).

The second insurance dataset is the state of Texas insurance dataset which is used by the Texas government to draw up a *Texas Liability Insurance Closed Claims Report*. The report provides a summary of claims involving bodily injuries from insurance companies. These claims were either settled in court or disposed of, and the insurer performed all the compensations and expense payments on the claim. There are two types of claims expressed in the dataset, long and short form. Short form focuses on claims on bodily injuries that are not expensive to settle. Long form relates to claims on bodily injuries that are very expensive and can be settled in most cases via a medical insurance company. In this dataset, we classify instances based on whether they have medical insurance cover as a risk analysis exercise provided there is missing data.

The Texas Insurance dataset consists of over 9000 instances, trimmed manually to 5446 instances by removing all the short form claims. For consistency, the dataset was separated into training and testing datasets, 4000 and 1446 instances respectively. Both the training and testing sets have missing values initially. Each set consists of a total of over 220 attributes initially, but the attributes were trimmed to 185 attributes. This was done by manually removing those attributes that were clearly not significant for the experiment, like the unique identities, dates as well the type of claim attributes. The class attribute here also has two values (0 to indicate no medical insurance and 1 to indicate that the claimer has medical insurance).

There are five levels of proportions of missingness on the testing dataset that were generated (10%, 25%, 30%, 40%, 50%). At each level, the missingness was arbitrarily generated across the entire dataset, then on half the attributes of the set. Therefore, in total, 12 testing datasets were created to test the strength of the Ripper algorithm using feature selection techniques.

### 3.2 PCA-RIP STRUCTURE

Figure 1illustrates the structure followed in improving the Ripper classification performance using the PCA as a feature selection technique. We refer to the structure as the PCA-Rip. From the figure, the original data [A] is given to the PCA. PCA reduces the dimensions of the data to give the output [T] expressed in equation (2).Attributes with eigenvalues > 1 were selected as a simple and effective approach to reduce the number of attributes. The Ripper algorithm builds a rule-based system using [T]. Once the Ripper algorithm is complete with learning the data, the PCA converts the data into its "original" data [A'] as expressed in equation (3). Data classification is performed using testing data.

The software tools used for this were Weka 3.6.2 library, C# 3.5 programming language and IKVM. Weka library has a built-in Principal Component analysis component. The component is used in conjunction with a Ranker search component to return the selected attributes in a chronological order from the most significant to the least significant attributes. IKVM is a software tool used to convert java code into C# code. The PCA-Rip illustrated in figure was built and tested using the C# programming language.

## 3.2 ARD-RIP STRUCTURE

Figure 2 illustrates the structure followed in improving the Ripper classification performance using the ARD as a feature selection technique. We refer to the structure as the ARD-Rip. Using the training data,each instance was expressed as an input vector **x** and supplied as original data as illustrated in figure 2. In some instances, the attributes for each instances were split into four groups where and supplied separately as input vectors $x_i$ as the original data to the Bayes artificial neural network.The reason for splitting the attributes is that the ARD performance is slow and memory intensive with highly dimensional datasets likes the Texas Insurance dataset.

The number of input units to the artificial neural network is equivalent to the size of the input vector and the number of hidden values is determined using trial and error. There is only one single output. The training is done over 1000 epochs, with the back-propagation algorithm as the learning algorithm. The output is evaluated using (2).The evidence framework re-evaluates the hyper-parameters and the prior probability of weights using equation (5) is calculated before supplying the input into the artificial neural network. Attributes with weight values < 0.01 we removed.

The ARD was built using Netlab and C# 3.5 programing language. Netlab has a built-in evidence automatic relevance determination model. A C# was designed to remove those attributes defined as irrelevant by the ARD before supplying to the Ripper algorithm.

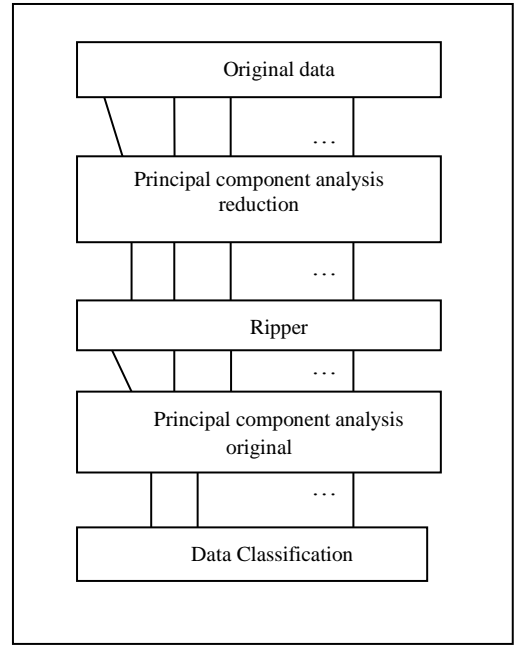

Figure 1: PCA-Rip structure

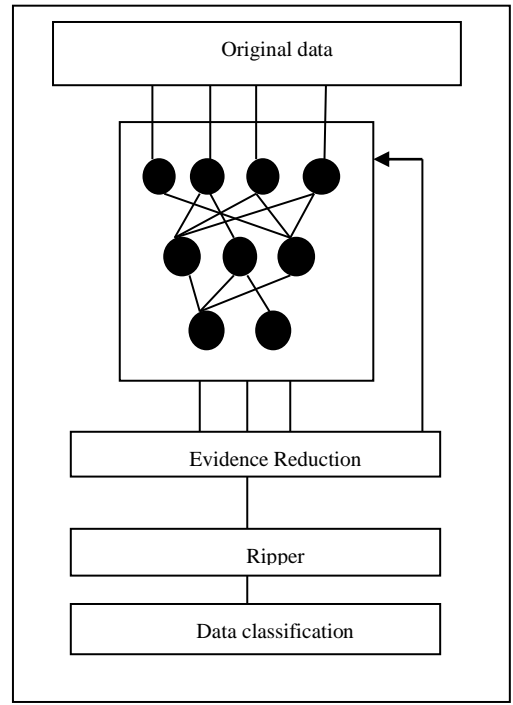

Figure 2: ARD-Rip structure

## 4 EXPERIMENT RESULTS

Table 1 illustrates the overall classification accuracy of the PCA-Rip and ARD-Rip compared to the Ripper algorithm. It can be noticed that both models performed well. PCA-Rip algorithm shows a significant improvement in accuracy compared to ARD-Dip. The reason for this is that the reduction technique used by the automatic relevance struggled to find relevant attributes in the datasets. A large number of $\alpha_c$ constants were either increasing or decreasing too quickly. Even in cases where the number of dimensions for the dataset was reduced significantly compared to PCA-Rip, ARD-Rip showed minimal improvement when compared to PCA-Rip.

Table 1: Overall classification accuracy.

|  | Accuracy (%) |
|---|---|
| Ripper | 87.85 |
| PCA-Rip | 91.96 |
| ARD-Rip | 88.87 |

Figure 3 shows the overall average performance of the algorithms in a chronological order of missingness in the dataset. From the figure, PCA-Rip performs better overall than the ARD-Rip. It shows more resilience and maintains high classification accuracies as the quality of data deteriorates. ARD-Rip struggles initially in performance compared to the Ripper. However, as the quality of data deteriorates, it shows more resilience and steadiness in performance.

Figure 4 shows the performance of all the models with half or all attributes having missing data. It is clear that the models perform better with missing data on half the attributes. Furthermore, the models show resilience and steadiness with increasing missingness on the data. With all or most attributes having missing data, the performance of the models decrease significantly (almost linearly) with little or no resilience.

The Ripper and the ARD-Rip models are the major contributor of this sharp decrease in performance. This is illustrated in figure 5. From the figure, the performance of the Ripper model is poor when half or all attributes have missing data. This was expected as explained earlier in the paper. The PCA-Rip model achieves high classification accuracies for datasets with half or all the attributes having missing data. ARD-Rip model performs as the Ripper with all attributes with missing data. However, it performs almost as well as the PCA-Rip model when half of the attributes have missing data.

The reason for this is that in some cases, the automatic relevance determination technique reduced the dimensions of a dataset significantly. This in return allowed the Ripper model to generate a rule set that managed to classify most instances from test data correctly.

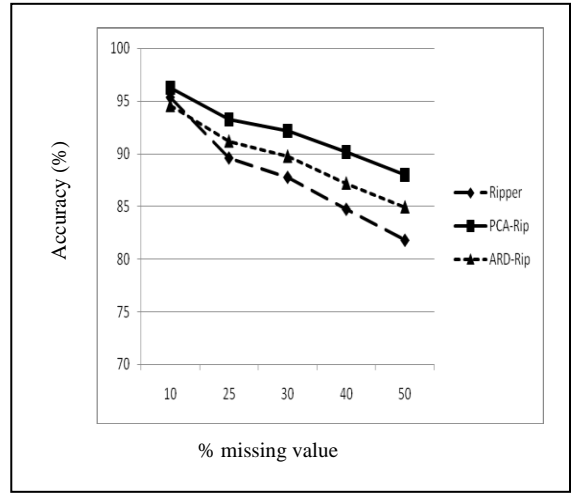

Figure 3: Overall average performance of the algorithms

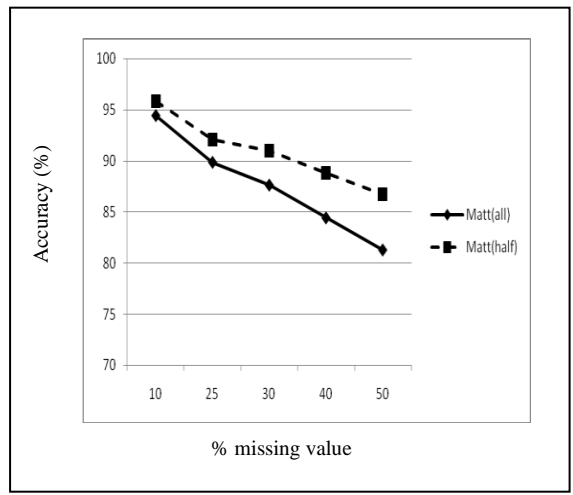

Figure 4: Overall average performance of the algorithms. Matt(all) represents all or most attributes with missing values. Matt(half) represents half the attributes with missing values.

## 5 CONCLUSION

A study on the PCA and ARD as feature selection techniques to improve the classification performance of the Ripper algorithm was conducted. Ripper showed to overall improvement when both techniques were used. With PCA technique, the

Ripper showed better results than with ARD. With PCA, the Ripper model achieved high classification accuracies and showed more resilience when data quality deteriorated. With ARD, the Ripper showed steadiness as the data quality deteriorated. However, it struggled to achieve high classification accuracies.

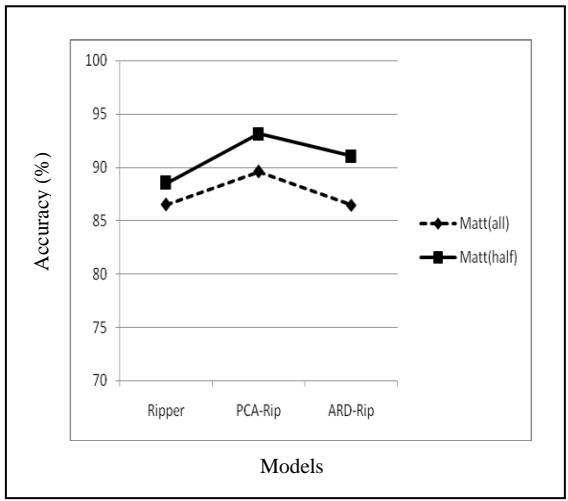

Figure 5: Overall performance of each model